%% file: main.tex
\documentclass[conference, letterpaper, final]{IEEEtran}

\IEEEoverridecommandlockouts

\input{includes}

\author{Andrey Solano$^{1,2}$, Arne Sieverling$^{1}$, Robert Gieselmann$^{2}$, Andreas Orthey$^{1,3}$ 
\thanks{$^{1}$Realtime Robotics Inc., Boston, MA, USA}%
\thanks{$^{2}$KTH Stockholm, Sweden}%
\thanks{$^{3}$TU Berlin, Germany}%
}
\title{\Huge Fast-dRRT*: Efficient Multi-Robot Motion Planning
using Decomposed Sampling-Based
Methods}
\title{\Huge Fast-dRRT*: Efficient Multi-Robot Motion Planning
using Swept-Volume Precomputation}
\title{\Huge Fast-dRRT*: Efficient Multi-Robot Motion Planning
for Automated Industrial Manufacturing}
\begin{document}

\maketitle

\pagenumbering{arabic}
\thispagestyle{plain}
\pagestyle{plain}

\input{src/00_abstract}
\input{src/01_introduction}
\input{src/02_relatedwork}
\input{src/03_background}
\input{src/04_method}
\input{src/05_evaluation}
\input{src/06_conclusion}

\bibliographystyle{IEEEtranS}
{
\balance
\small
\bibliography{IEEEabrv, bib/general}
}
\end{document}

%% file: includes.tex
\usepackage{xcolor} 

\usepackage{graphicx} 
\usepackage{balance} 
\usepackage{hyperref} 
\usepackage{amsmath} 
\usepackage{subcaption} 
\usepackage{multirow} 
\usepackage{tabularx}
\usepackage[font=small,labelfont=bf]{caption}

\usepackage[left=19.1mm,right=19.1mm,top=25.4mm,bottom=19.1mm]{geometry}

\def\X{X}
\def\Xi{X_i}

\def\xstart{x_{\text{start}}}
\def\xistart{\ensuremath{x_{i, \text{start}}}}

\usepackage{algorithm}
\usepackage{algpseudocode} 
\usepackage{booktabs} 

\makeatletter

\algnewcommand{\algorithmicparams}{\textbf{Parameters:}}
\algnewcommand\Params{\item[\algorithmicparams]}
\algnewcommand{\algorithmiccontinue}{\textbf{continue}}
\algnewcommand{\algorithmicbreak}{\textbf{break}}
\algnewcommand{\BREAK}{\algorithmicbreak}
\algnewcommand{\algorithmicor}{\textbf{ or }}
\algnewcommand{\OR}{\algorithmicor}
\algnewcommand{\algorithmicand}{\textbf{ and }}
\algnewcommand{\AND}{\algorithmicand}
\algnewcommand{\CONTINUE}{\algorithmiccontinue}
\algnewcommand\True{\textbf{true}\xspace}
\algnewcommand\False{\textbf{false}\xspace}
\makeatother

\makeatletter
\renewcommand{\midrule}{\kern2pt\hrule\kern2pt} 
\newcommand{\algcaption}[2][]{%
  \refstepcounter{algorithm}%
\textbf{{\raggedright\fname@algorithm~\thealgorithm}}\ #2\par 
  \midrule
}
\makeatother

\makeatletter
\newcommand*{\algrule}[1][\algorithmicindent]{\makebox[#1][l]{\hspace*{.5em}\vrule height .75\baselineskip depth .25\baselineskip}}%

\newcount\ALG@printindent@tempcnta
\def\ALG@printindent{%
    \ifnum \theALG@nested>0
    \ifx\ALG@text\ALG@x@notext
    \addvspace{-3pt}
    \else
    \unskip
    \ALG@printindent@tempcnta=1
    \loop
    \algrule[\csname ALG@ind@\the\ALG@printindent@tempcnta\endcsname]%
    \advance \ALG@printindent@tempcnta 1
    \ifnum \ALG@printindent@tempcnta<\numexpr\theALG@nested+1\relax
    \repeat
    \fi
    \fi
}%
\usepackage{etoolbox}
\patchcmd{\ALG@doentity}{\noindent\hskip\ALG@tlm}{\ALG@printindent}{}{\errmessage{failed to patch}}
\makeatother

\makeatletter
\algnewcommand{\LineComment}[1]{\Statex \hskip\ALG@thistlm \(\triangleright\) #1}
\makeatother

\def\vstart{v_{\text{start}}}
\def\vnear{v_{\text{near}}}
\def\vnew{v_{\text{new}}}
\def\vlast{v_{\text{last}}}
\def\xrand{x_{\text{rand}}}
\def\xgoal{x_{\text{goal}}}
\def\vgoal{v_{\text{goal}}}
\def\costv{\text{cost}_{v}}
\def\costvnew{\text{cost}_{v\text{new}}}
\def\tree{T}
\def\path{\pi}
\def\pathbest{\path_{\text{best}}}

\def\dvoxel{\ensuremath{d_{\text{voxel}}}}

%% file: src/00_abstract.tex
\begin{abstract}
We present Fast-dRRT*, a sampling-based multi-robot planner, for real-time industrial automation scenarios.
Fast-dRRT* builds upon the discrete rapidly-exploring random tree (dRRT*) planner, and extends dRRT* by using pre-computed swept volumes for eﬀicient collision
detection, deadlock avoidance for partial multi-robot problems, and a simplified rewiring strategy.
We evaluate Fast-dRRT* on five challenging multi-robot scenarios using two to four industrial robot arms from various manufacturers. The scenarios comprise situations involving deadlocks, narrow passages, and close proximity tasks. The results are compared against dRRT*, and show Fast-dRRT* to outperform dRRT* by up to 94\% in terms of finding solutions within given time limits, while only sacrificing up to 35\% on initial solution cost.
Furthermore, Fast-dRRT* demonstrates resilience against noise in target configurations, and is able to solve challenging welding, and pick and place tasks with reduced computational time.
This makes Fast-dRRT* a promising option for real-time motion planning in industrial automation.
\end{abstract}

%% file: src/01_introduction.tex
\section{Introduction}

Many complex industrial tasks like welding, palletizing, or pick-and-place often require multi-robot motion planning~\cite{marvel_multi-robot_2019}. 
Multi-robot planning problems can be solved using sampling-based algorithms~\cite{lavalle_planning_2006}, which randomly explore the configuration space. Planner like RRT*~\cite{karaman_anytime_2011} are both probabilistically complete and asymptotically optimal.
However, they typically suffer from the curse of dimensionality, and cannot solve problems involving high-dimensional multi-robot configuration spaces. 

Decomposition methods~\cite{sharon_conflict-based_2012-1, shome_drrt_2020, ferner_odrm_2013} tackle this problem by decomposing the configuration space, solving for individual robots, and then putting those solutions back together. 
\input{images/cover}
A prominent planner is dRRT*~\cite{shome_drrt_2020}, which can solve decomposed configuration spaces for arbitrary multi-robot planning scenarios---while being probabilistically complete, and asymptotically optimal.

However, dRRT* is not directly optimized for industrial manufacturing. 
First, collision detection is computed on-the-fly, thereby wasting valuable computation time. 
Second, dRRT* does not support deadlock avoidance. 
A deadlock describes a scenario where a robot without a goal obstructs the movement of another, preventing the latter from reaching its destination. Third, dRRT* does not optimize for finding a quick first solution, which is often critical in industrial settings. 

To address those problems, we develop an extension of dRRT*, which we call Fast-dRRT*.
Fast-dRRT* focuses on achieving high-quality collision-free paths in the shortest possible time.
This is achieved by using pre-computed swept volumes~\cite{murray_robot_2016} to optimize the collision detection during the search phase. 
Fig.~\ref{fig:AutomotiveAutomation} shows the use of swept volumes in a welding task, where three robots with an attached welding gun are working on parts of a car chassis. The left robot moves inside a pre-computed swept volume, and would collide with the idle robot in the middle.   
Fast-dRRT* resolves such situations using deadlock avoidance which move blocking robots out of the way, and simplifies the dRRT* code to reach a first initial solution quicker.
As in dRRT*, Fast-dRRT* is asymptotically optimal given infinite time. To summarize, our contributions are:

\input{images/system}
\begin{itemize}
    \item We develop the Fast-dRRT* planner, which avoids costly rewiring cycles of dRRT* to improve runtime.
    \item We extend both Fast-dRRT*, and dRRT* to partial multi-robot problems using a deadlock avoidance strategy, and use pre-computed swept volumes for efficient collision checking.
    \item We evaluate both Fast-dRRT*, and dRRT* on five multi-robot scenarios using industrial manipulators from various manufacturers. 
\end{itemize}

%% file: images/cover.tex
\begin{figure}[t]
  \begin{center}
    \includegraphics[width=\linewidth]{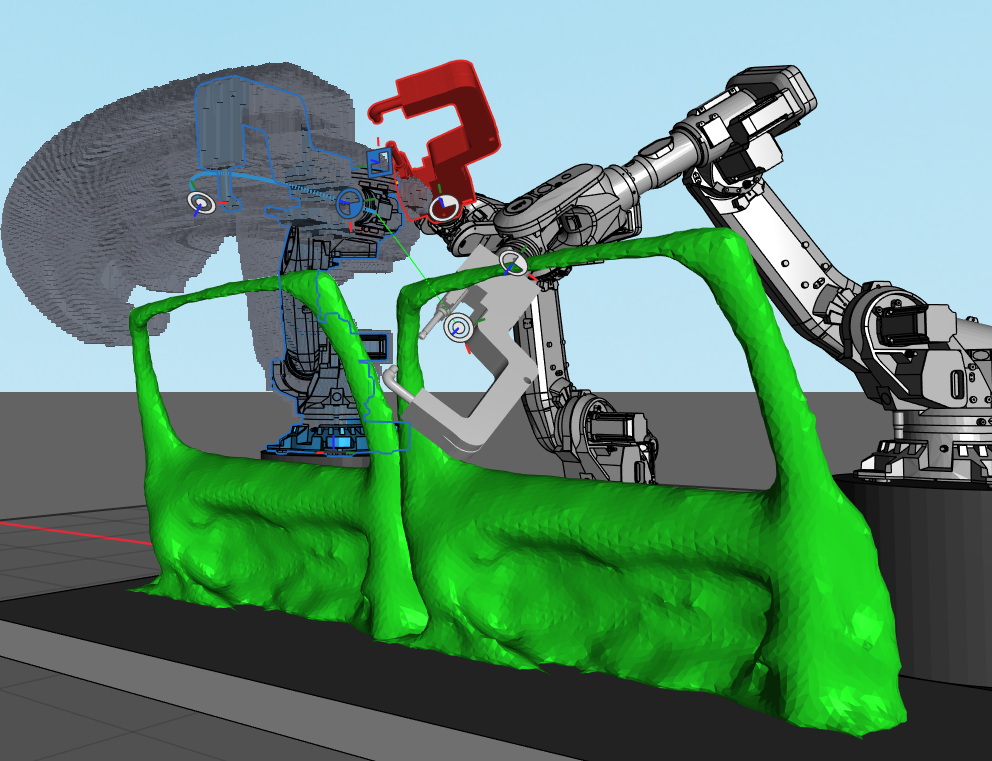}
  \end{center}
  \caption{Multiple robots collaborating on a car chassis welding task. The left robot moves along a path (visualized by the grey swept volume), and would collide with the middle robot (red links) which is currently idle. Fast-dRRT* can move idle robots out of the way, exploit pre-computed swept volumes, and avoid costly rewiring cycles to efficiently solve multi-robot planning problems.\label{fig:AutomotiveAutomation}}
  \vspace*{-0.5cm}
\end{figure}

%% file: images/system.tex
\begin{figure*}[t]
  \begin{center}
    \includegraphics[width=\linewidth]{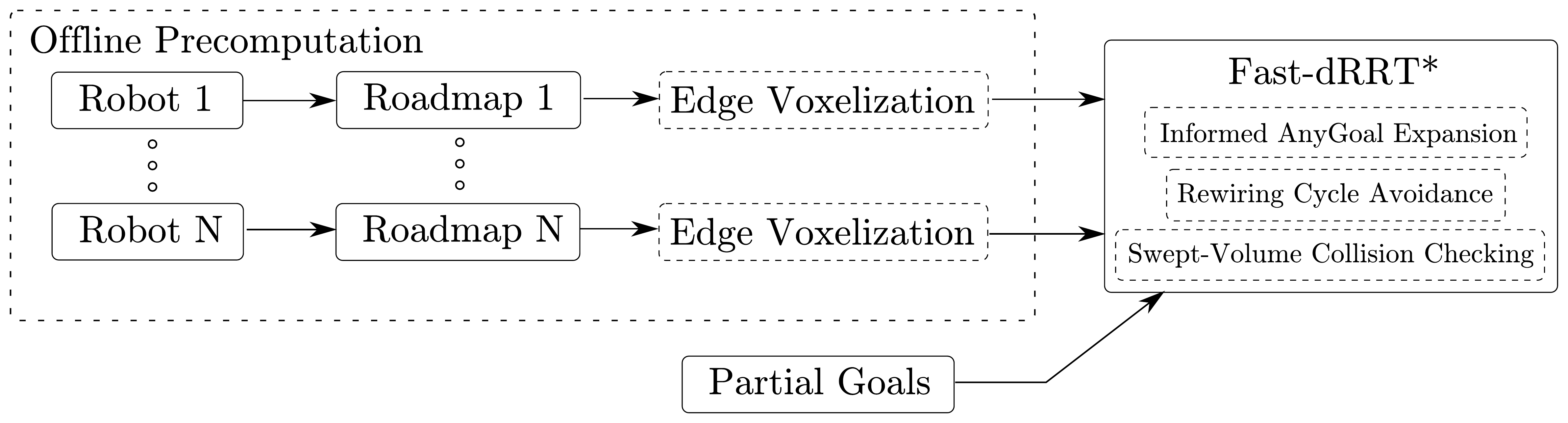}
  \end{center}
  \caption{Overview about Fast-dRRT* including all precomputation steps. We solve multi-robot scenarios for $N$ robots by computing $N$ roadmaps for each individual robot (Sec.~\ref{sec:roadmap_generation}) For each roadmap, we approximate each edge swept volume using edge voxelization (Sec.~\ref{sec:edge_voxelization}) Those roadmaps with precomputed swept-volumes are then used together with partial goals as input to Fast-dRRT*. Inside Fast-dRRT* we use a custom informed expansion step (Sec.~\ref{sec:informed_expansion}) to avoid deadlock situations, use avoidance of rewiring cycles (Sec.~\ref{sec:avoidance_rewiring}) to quickly find a first initial solution, and use the pre-computed swept volumes for fast collision checking (Sec.~\ref{sec:collison_checking}).\label{fig:system}}
\end{figure*}

%% file: src/02_relatedwork.tex
\section{Related Work}

Multi-robot motion planning problems often benefit from reduction methods. In the case of homogeneous robot teams~\cite{yan_survey_2013}, i.e. where all robot models are equivalent, we can reduce the problem to the pebbles-on-a-graph problem~\cite{kornhauser1984coordinating, papadimitriou1994motion, yu2016optimal}. This methodology has been widely employed, for example for swarms of drones~\cite{honig2018trajectory}.  
Our method differs by allowing heterogeneous robot scenarios, since fixed robots at different positions have different configuration spaces, and since we aim to support different robots from different manufacturers. 

For heterogeneous multi-robot teams, planning can often be reduced using either prioritization or decomposition. 
In prioritization, a solution is computed for a single robot, whose motion is then used as a constraint for the second robot~\cite{erdmann_multiple_1987, van_den_berg_prioritized_2005, van_den_berg_centralized_2009}. 
Such a prioritization approach is often efficient~\cite{saha_implan_2016, Orthey2021WAFR}, but requires an ordering of the robots. 
While there are asymptotically-optimal versions of prioritization~\cite{Orthey2020IJRR}, inefficient orderings can sometimes lead to excessive backtracking. 
Our method differs by focusing on decompositions of the configuration space, since this not only removes the ordering of robots, but also allows us to pre-compute swept volumes, and thereby improve execution time significantly. 

Finally, multi-robot planning problems can often be reduced using decomposition methods~\cite{wagner_subdimensional_2015, solovey_finding_2016}. In a decomposition method, each robot computes their own roadmap ignoring all other robots. To solve the complete problem, the roadmaps are combined in an implicit graph which can be searched through. 

An important planner for implicit graphs is the M* method~\cite{wagner_m_2011, wagner_subdimensional_2015}. M* extends the A* graph search planner~\cite{hart_formal_1968} to multi-robot scenarios by using the shortest paths from the individual roadmaps as admissible heuristics. M* has the same optimality guarantees as A*, and is therefore a good candidate to explore implicit graphs. However, M* can lead to high planning times since it requires a best cost estimate for all neighbors of a vertex, which can often be costly in high-dimensional spaces~\cite{solovey_finding_2016}.

An improvement upon M* for euclidean state spaces is the discrete rapidly-exploring random tree (dRRT) planner~\cite{solovey_finding_2016}. Instead of building a search graph and computing costs for neighborhood vertices directly, dRRT employs a tree and a proxy method for neighborhood costs. This proxy method is based on an oracle~\cite{solovey_finding_2016}, which selects the neighboring vertex that minimizes the angle between a line from the current vertex to the neighboring vertex and a line from the current vertex to a random sampled state. This speeds up the search significantly~\cite{solovey_finding_2016}. 

Several methods have been developed which extend dRRT to make it asymptotically optimal. The asymptotically-optimal dRRT (ao-dRRT)~\cite{shome_drrt_2020} adds both a rewiring method if an edge is already in the tree, and checks if a new vertex improves the current path cost. A further extension of dRRT is the dRRT* planner~\cite{shome_drrt_2020}, which replaces the oracle with an informed expansion strategy. This informed expansion strategy exploits the optimal paths from the individual state spaces to steer the search towards low-cost paths. 

Our method Fast-dRRT* improves upon dRRT* by using pre-computed swept volumes for efficient collision checking, by adding support for partial multi-robot planning problems, and by replacing the rewiring cycle of dRRT*, which reduces the computational time to find a first solution.

%% file: src/03_background.tex
\section{Multi-Robot Motion Planning}

Let $\X = \X_1 \times \cdots \times \X_N$ be the state space of $N$ robots $R_i$, whereby $\Xi$ is the individual state space of robot $R_i$, and $\times$ is the Cartesian product of the individual state spaces.

The multi-robot motion planning problem is to find a collision-free path, $\pi: [0,1] \rightarrow \X$, from a start state $\xstart$ to a goal region $\X_G$, whereby each robot has its own start state $\xistart$, and own goal region $\X_{i, G}$. 
A \emph{partial multi-robot problem} is defined by a goal region which is only defined for a subset of robots $R_{J}$, i.e. there are robots without a specific goal, which is a typical situation in manufacturing scenarios. We model such problems by providing an \emph{any goal} region, i.e. $\X_{i, G} = \X_i$. 
Note that a partial multi-robot problem is different from the identity problem, where start and goal are equivalent for some robots. 
As an example, consider a robot which blocks a goal position of another robot, and has to move away to solve the problem. This cannot explicitly be solved by dRRT*, since dRRT* relies on an explicit definition of goal states for each robot.

%% file: src/04_method.tex
\input{algorithms/fast_drrt}
\input{algorithms/update_path}

\section{Fast Discrete-RRT*\label{sec:fast-dRRTStar_meth}}

Fast-dRRT* is an asymptotically-optimal multi-robot planner which is designed as an adaptation of dRRT*~\cite{shome_drrt_2020} (See Fig.~\ref{fig:system} for an overview). While Fast-dRRT* has a similar structure as the original dRRT*~\cite{shome_drrt_2020}, it differs by using a different roadmap generation (Sec.~\ref{sec:roadmap_generation}), pre-computed swept volumes for collision checking (Sec.~\ref{sec:edge_voxelization}), an informed any-goal expansion for deadlock avoidance (Sec.~\ref{sec:informed_expansion}), and no rewiring cycle to get quick first solutions (Sec.~\ref{sec:avoidance_rewiring}).

The complete Fast-dRRT* method is shown in Alg.~\ref{alg:fastdrrt}. Fast-dRRT* starts by initializing a tree (Line 1), and a vertex $\vlast$ to keep track of the last vertex added to the tree (Line 2). Then, a while loop is started until a planner terminate condition (PTC) is reached (Line 3). The PTC can be a maximum number of iterations, finding a first solution, a cost convergence, a time limit, or a combination of them. If $\vlast$ is not empty (Line 4) it means that we are in the first iteration or that the previous expand iteration was successful, here we expand towards the goal (Line 5) from the last added vertex (Line 6). If $\vlast$ is empty, we sample a random state (Line 8), and expand the tree from the nearest vertex (Line 9). An informed expansion (Line 11), which is further detailed below, finds $\vnew$. Its cost is given by the current cost of $\vnear$ in the tree plus the cost to move from $\vnear$ to $\vnew$ (Line 12). We then create a new edge (Line 13), and clear the content of $\vlast$ (Line 14). If the edge is invalid or the cost is above the current best cost (Line 15-16), we terminate the iteration and continue. If the new edge is already an element of the tree (Line 18), we check if the cost of $\vnew$ via $\vnear$ is lower than the current cost of the vertex in the tree (Line 19-20), in which case we rewire the tree (Line 23), similar to the rewiring step in RRT*~\cite{karaman_anytime_2011}. If the edge is not yet in the tree, we add it to the tree (Line 25). Finally, we update the path as detailed in Alg.~\ref{alg:updatepath} (Line 27), and set the value for $\vlast$ (Line 28) before the next iteration is started.

\subsection{Roadmap Generation\label{sec:roadmap_generation}}

The original dRRT*~\cite{shome_drrt_2020} advocates a sampling-based method such as PRM~\cite{kavraki_probabilistic_1996}, to generate a fixed-size roadmap on the individual state spaces of each robot. However, in industrial planning problems, we often face situations, where robots have multiple goals (targets) which are concentrated in a smaller subspace. We therefore opted for a roadmap generation which uses best-cost paths between targets to avoid including nodes that are too far away from the interest area. This is achieved by using RRT*~\cite{karaman_sampling-based_2011}, or RRT-Connect~\cite{kuffner_rrt-connect_2000} for each target pair, and store all those paths in the roadmap. This creates a simple, sparse roadmap which is similar to the experience graph~\cite{phillips2012graphs}. Note that Fast-dRRT* can operate on any roadmap, and the generation of the underlying roadmap is scenario specific. 

\subsection{Swept Volumes Computation\label{sec:edge_voxelization}}

\begin{figure}
    \centering
    \begin{subfigure}{0.48\linewidth}
        \includegraphics[width=\linewidth, height=0.8\linewidth]{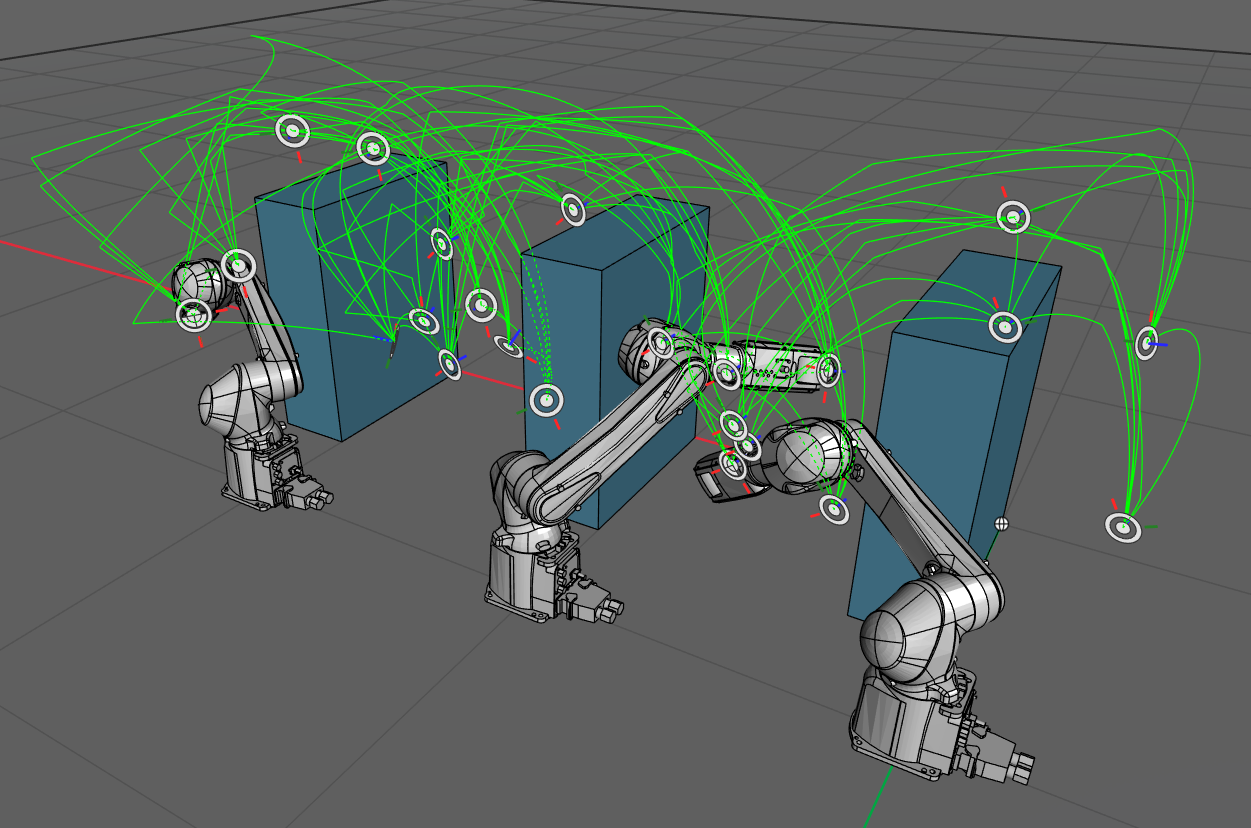}
    \caption{Roadmap generation using lowest-cost paths between targets.\label{fig:roadmap_generation}}
    \end{subfigure}
    \begin{subfigure}{0.48\linewidth}
        \includegraphics[width=\linewidth, height=0.8\linewidth]{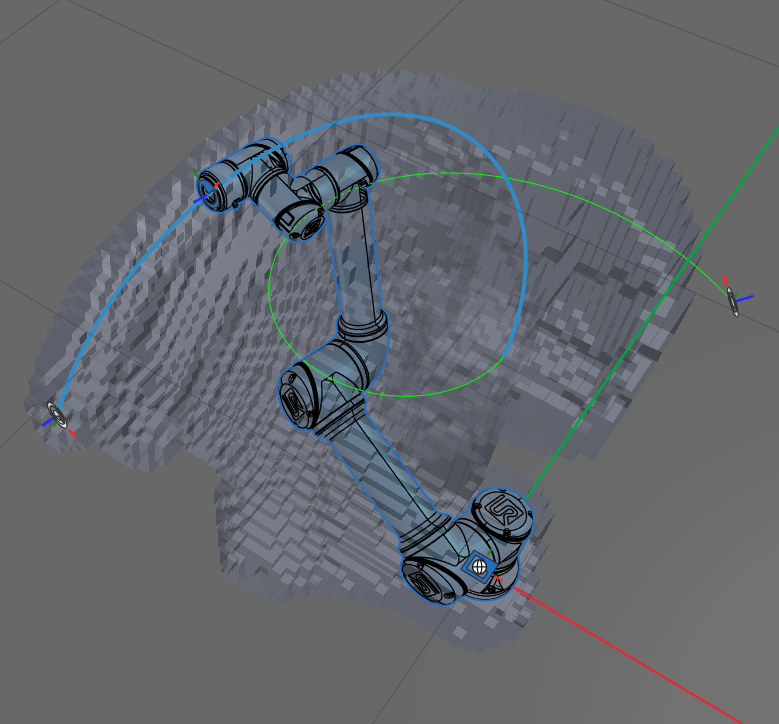}
        \caption{Edge voxelization to approximate the swept volume of a robot.\label{fig:edge_voxelization}}
    \end{subfigure}
    \caption{Different stages for the pre-computations of Fast-dRRT*. }
\end{figure}

Once we computed a roadmap for each individual robot, we pre-compute swept volumes for each edge in the roadmap using edge voxelization (see Fig.~\ref{fig:edge_voxelization}). Edge voxelization works by approximating the swept volumes using voxels of length $\dvoxel$. We first define a reachable bounding box, which contains all geometries of the robot at every possible configuration. We then discretize every edge into a set of configurations, each configuration $\delta$-spaced apart along the edge. For each configuration, we compute the forward kinematics, and check for each voxel in the reachable bounding box if it collides with the geometry. All the colliding voxels are stored. Once this is done for all configurations along an edge, we remove overlapping voxels and store the resulting voxel geometry in the edge for collision checking.

\subsection{Collision Checking\label{sec:collison_checking}}

The online robot-to-robot collision checking leverages the pre-computed swept volumes for each edge in the individual roadmaps. An edge in the combined graph is collision free if there is no intersection between the swept volumes of the edges that compose it.
Given that we represent the swept volumes using their voxelized approximation, the collision detection is reduced to determine if the set of intersecting voxels is empty or not.

\input{algorithms/informed_expansion}

\subsection{Informed AnyGoal Expansion\label{sec:informed_expansion}}

The informed expansion in Fast-dRRT* happens similarly as in dRRT* \cite{shome_drrt_2020}. However, contrary to dRRT*, we tackle partial problems, where one or more robots have no specified goal, but could impede the others to reach their goals. It is critical to account for those robots, because they would have no incentive to move. We model those undefined goals using an \emph{any-goal}, which is always satisfied.

The adapted informed expansion keeps the robots without a defined goal at their place and only moves them after an iteration fails to expand the tree. This is modelled in the informed expansion method in Alg. \ref{alg:informedexpansion}. 
In this method, we iterate over all individual robot state spaces (Line 1). Depending if $\vlast$ is empty (Line 2), we move all robots to a random neighbor, including the any-goal robots (Line 3). Otherwise, we try to keep moving towards the goal, without moving the any-goal robots (Line 5-9).

\subsection{Avoidance of Rewiring Cycle\label{sec:avoidance_rewiring}}

In dRRT*, an additional rewiring cycle is used to minimize the distance along the tree~\cite{shome_drrt_2020}. This rewiring cycle adds to Alg.~\ref{alg:fastdrrt} an additional step between Line 11 and Line 12. In this additional step, a neighboring set of $\vnew$ is computed together with a best cost-to-go neighbor. This requires a costly computation of adjacency in the implicit Graph $G$. The best neighbor is then used to make an additional rewire step. This ensures that the tree always updates the best cost-to-go estimates.
However, in our scenarios, it is often important to quickly find a first solution. We therefore make the trade-off of removing this rewiring cycle, in order to have faster runtime at higher initial cost. 

%% file: algorithms/fast_drrt.tex

\begin{algorithm}[t]
\algcaption{Fast-dRRT*}
\begin{algorithmic}[1]
    \Require State space $\X$, Implicit graph $G$, vertex $\vstart$
    \Ensure Path $\pathbest$
    \State $\tree \gets \Call{InitializeTree}{\vstart}$ 
    \State ${\vlast} \gets {\vstart}$
    \While{$\neg$ PTC}
        \If{$\vlast \neq \emptyset$}
            \State $\xrand \gets \xgoal$
            \State $\vnear \gets \vlast$
        \Else
            \State $\xrand \gets \Call{Random}{\X}$
            \State $\vnear \gets \Call{Nearest}{\tree, \xrand}$
        \EndIf
        \State $\vnew \gets \Call{InformedAnyGoalExpansion}{\vnear, \xrand}$
        \State $\costvnew \gets \Call{Cost}{\vnear} + \Call{Cost}{\vnear, \vnew}$ 
        \State $e \gets \Call{Edge}{\vnear, \vnew}$
        \State $\vlast \gets \emptyset$
        \If{$\neg \Call{IsValid}{e}\ \Call{Or}{}\ \costvnew > \Call{Cost}{\pathbest} $}
            \State \CONTINUE
        \EndIf
        \If{$\Call{IsElementOf}{\vnew, \tree}$}
            \State $\costv \gets \Call{Cost}{\vnew, \tree}$ 
            \If{$\costvnew \geq \costv$}
                \State \CONTINUE
            \EndIf
            \State $\Call{Rewire}{\vnear, \vnew, \tree}$
        \Else
            \State $\Call{AddEdgeToTree}{e, \tree}$
        \EndIf
        \State $\pathbest \gets \Call{UpdatePath}{\pathbest, \tree, \vnew}$
        \State $\vlast \gets \vnew$
    \EndWhile
    \State \Return $\pathbest$
\end{algorithmic}\label{alg:fastdrrt}
\end{algorithm}

%% file: algorithms/update_path.tex
\begin{algorithm}[t]
\algcaption{UpdatePath}
\begin{algorithmic}[1]
    \Require Path $\pathbest$, $\tree$, $\vnew$
    \Ensure Path $\pathbest$
    \If{$\Call{GoalIsSatisfied}{T}$}
        \State $\vstart \gets \Call{GetStartVertex}{\tree}$
        \State $\vgoal \gets \Call{GetGoalVertex}{\tree}$
        \State $\path \gets \Call{Path}{\tree, \vstart, \vgoal}$
        \If{$\Call{Cost}{\path} < \Call{Cost}{\pathbest}$}
            \State $\pathbest \gets \path$
        \EndIf
    \EndIf    
    \State \Return $\pathbest$
\end{algorithmic}\label{alg:updatepath}
\end{algorithm}

%% file: algorithms/informed_expansion.tex
\begin{algorithm}[t]
\algcaption{InformedAnyGoalExpansion}
\begin{algorithmic}[1]
    \Require Vertex $\vlast$, Vertex $\vnear$, State $\xrand$ 
    \Ensure Vertex $\vnew$
    \For{$i$ in $\{1,\ldots,N\}$}
        \If{$\vlast = \emptyset$}
            \State $\vnew^i \gets \Call{RandomNeighbor}{\vnear^i}$
        \Else
            \If{$\Call{IsAnyGoal}{i}$}
                \State $\vnew^i \gets \vnear^i$
            \Else
                \State $\vnew^i \gets \Call{NextElementOnPath}{\path^i, \vnear^i}$
            \EndIf
        \EndIf
    \EndFor    
    \State \Return $\vnew$
\end{algorithmic}\label{alg:informedexpansion}
\end{algorithm}

%% file: src/05_evaluation.tex
\input{images/evaluations/figure_scenarios}
\input{images/evaluations/figure_evaluation_table}

\section{Evaluation}

We evaluate Fast-dRRT* on $5$ realistic industrial scenarios with $2$ to $4$ robots. All five scenarios are shown in Fig.~\ref{fig:scenarios}. In each experiment, we compare the performance of Fast-dRRT* to dRRT*~\cite{shome_drrt_2020} on both runtime and cost convergence. For a fair comparison, we apply both swept-volume pre-computation, and the new informed any-goal expansion strategy to dRRT*. This version of dRRT* differs only by the additional rewiring cycle compared to Fast-dRRT* (see Sec.~\ref{sec:avoidance_rewiring}). 

Concerning hardware, we use a PC with an Intel i7 eight core CPU, $64$GB RAM, and a Quadro P1000 graphics card.
The edge voxelization uses the parameter $\dvoxel=1$cm. 

For each scenario, we run two experiments. First, the robots have to reach a sequence of goal configurations (targets). For each segment in the sequence, we repeat our evaluations $100$ times, and store the mean time to the first solution, the worst time to the first solution, the mean cost at the first solution, and the mean cost after $100$ms. The cost used is the path length of a solution. The results are shown in Table~\ref{table:results}.
We then run a second experiment, where all targets are randomly perturbed in each run to check if the methods are robust against noise. Times and costs are presented as histograms, to give more detailed insights into the planner results. Those results are shown in Fig.~\ref{fig:time_evaluation} (time), and Fig.~\ref{fig:cost_evaluation} (cost), respectively. 

\subsection{Deadlock Aisles}

The first scenario contains three Kawasaki RS007L robots~\cite{kawasaki_robotics_rs007l_nodate}, whereby each robot has to reach target locations both to the left and to the right of a block. Each robot has $7$ targets it has to reach. The results in Table~\ref{table:results} show that Fast-dRRT* finds a solution $22$\% faster than dRRT*, while the cost is on average $17$\% higher.
After 100ms of execution the
mean cost of the solutions of Fast-dRRT* are $25$\% higher.

The second experiment shows similar results. As shown in Fig.~\ref{fig:time_evaluation}, Fast-dRRT* finds a first solution for the perturbed targets always under $10$ms with most results below $5$ms. dRRT* performs less well with all first solutions below $20$ms. The results for costs in Fig.~\ref{fig:cost_evaluation} are similar, but Fast-dRRT* has more outlier with higher cost. 

\subsection{Deadlock Table}

This scenario consists of four ABB IRB 2600 robots arranged symmetrically around a cylindrical obstacle, as shown in Figure \ref{fig:scenarios}. The goal configurations of each robot are identical. Fast-dRRT* finds a solution $63$\% faster, while having an initial average solution cost which is $9$\% higher. After $100$ms, the cost is $25$\% higher for solutions found so far.

The second experiment shows that Fast-dRRT* finds all first solutions in $100$ms with one exception at $150$ms (Fig.~\ref{fig:time_evaluation}), while dRRT* has a wider spread of results up to $300$ms. Costs are comparable, with Fast-dRRT* having a slightly higher first cost. 

\subsection{Narrow Passage}

This scenario comprises three Mitsubishi RV-4FRL \cite{mitsubishi_electric_corporation_rv-4frl-d_nodate} robots, two of which are mounted on columns to increase their height. Fig.~\ref{fig:scenarios} presents the arrangement of the robots. 
Fast-dRRT* outperforms dRRT* by $94$\% on the initial time to reach the first solution, while having an initial cost which is $35$\% higher than dRRT*. 

The time in the second experiment shows almost all results for Fast-dRRT* being below $30$ms, while dRRT* has a wider spread of results up to $300$ms. The cost histogram, however, shows slightly better results for dRRT*.
While the initial solution times in the second experiment are comparable, dRRT* has more outlier above $400$ms (Fig.~\ref{fig:time_evaluation}). 
\input{images/evaluations/figure_evaluation_time}
\input{images/evaluations/figure_evaluation_cost}
\subsection{Welding}

In this welding scenario, three ABB IRB 6700 industrial robots are mounted on separate columns. Each robot is equipped with a welding gun (Fig.~\ref{fig:scenarios}). The target welding points are taken from a robot cell used in an industrial car manufacturing task. 

The task of the robots consist of welding a set of assigned points on the car chassis. The results indicate that Fast-dRRT* outperforms dRRT* by $33$\% in initial solution time. Concerning cost, dRRT* obtains a $22$\% better initial solution cost. 
The second experiment shows that all results for Fast-dRRT* are below $400$ms, while dRRT* has a wider spread with outliers above $600$ms. 

\subsection{Pick and Place}

Two Yaskawa Motoman HC10 \cite{yaskawa_america_inc_yaskawa_nodate} robotic arms are mounted on a column in front of a table with nine colored boxes. The workspace setup is illustrated in Figure \ref{fig:scenarios}. The results show a $15$\% lead of Fast-dRRT* over dRRT* in initial solution time, and a $20$\% higher initial cost. 

The second experiment shows similar initial solution times, but with larger frequency of results below $10$ms for Fast-dRRT*. Fast-dRRT*, however, has slightly more outliers above $150$ms, and a higher overall cost.

%% file: images/evaluations/figure_scenarios.tex
\def\subfigWidth{0.19\textwidth}
\def\envWidth{\textwidth}
\def\envHeight{\textwidth}

\begin{figure*}
    \centering
    \begin{subfigure}{\subfigWidth}
        \includegraphics[width=\envWidth, height=\envHeight]{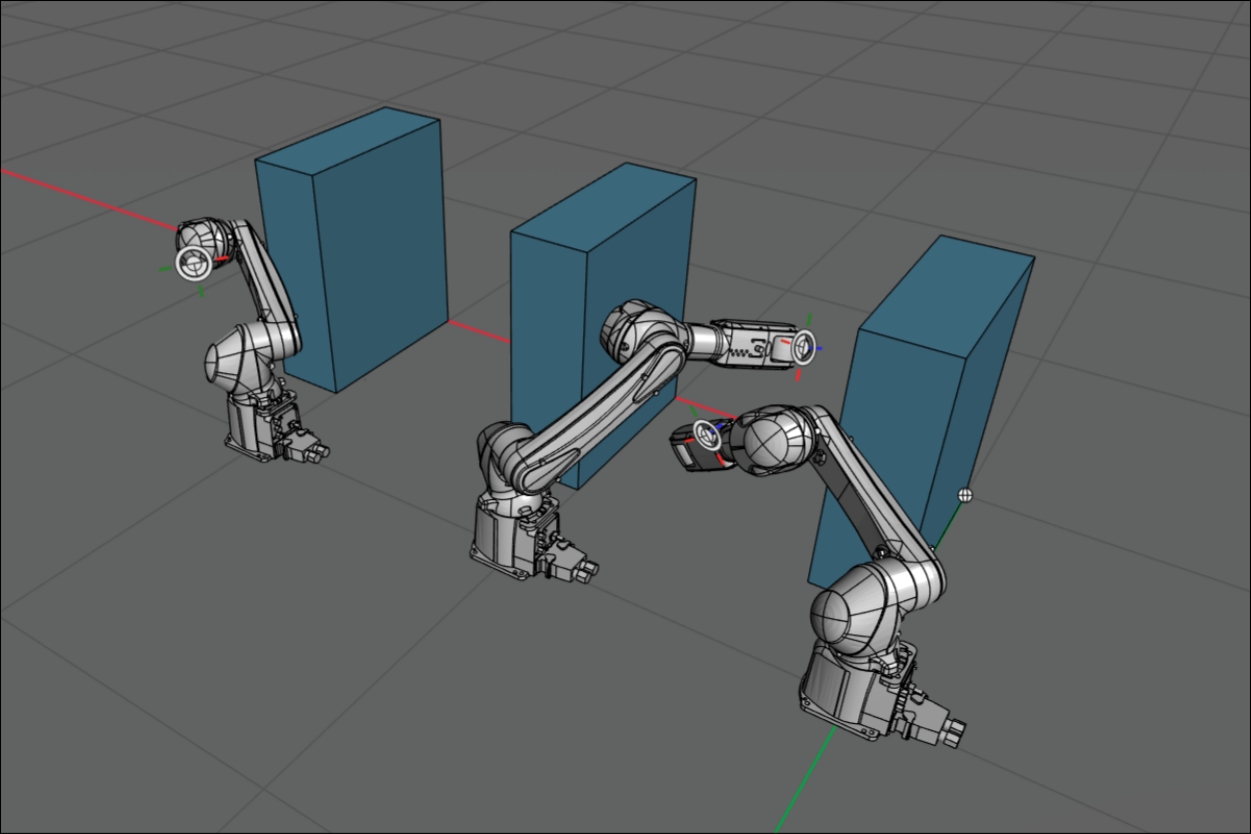}
        \caption{Deadlock Aisle}
    \end{subfigure}
    \begin{subfigure}{\subfigWidth}
        \includegraphics[width=\envWidth, height=\envHeight]{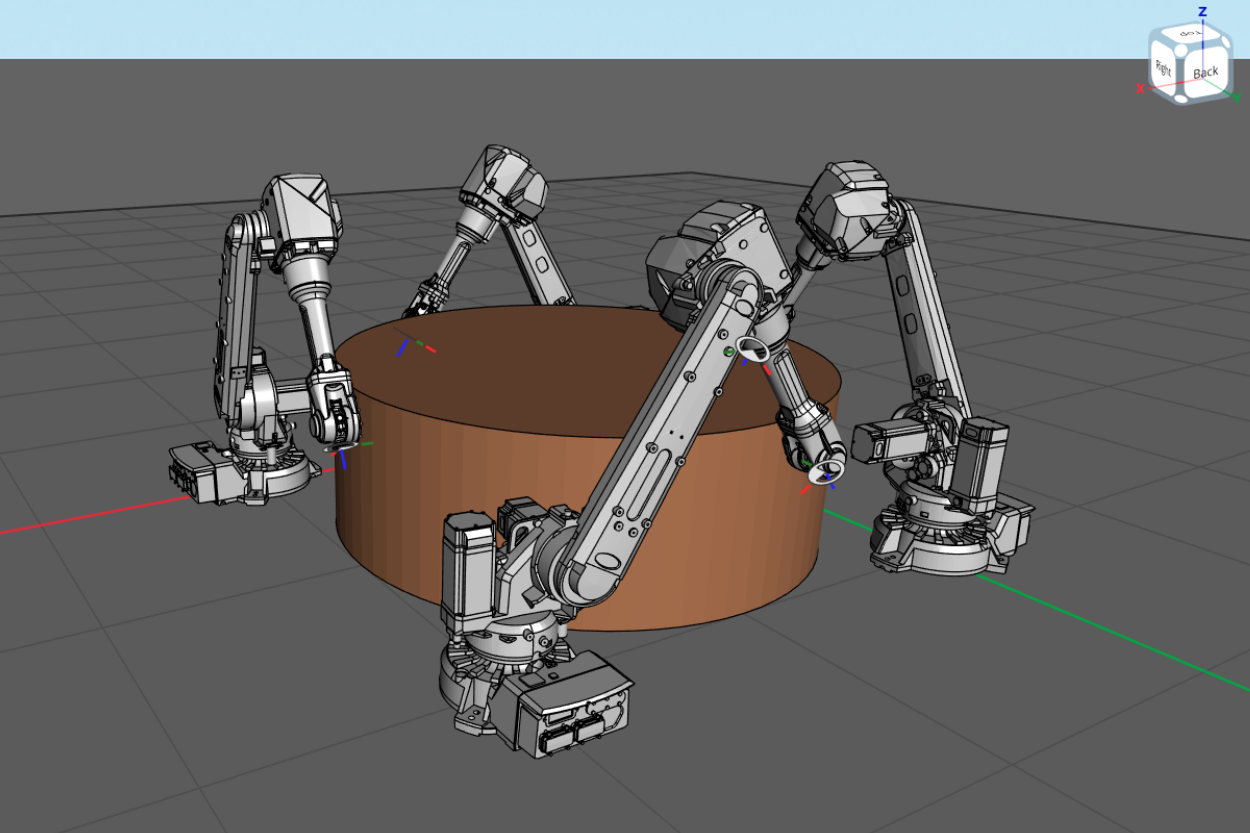}
        \caption{Deadlock Table}
    \end{subfigure}    
    \begin{subfigure}{\subfigWidth}
        \includegraphics[width=\envWidth, height=\envHeight]{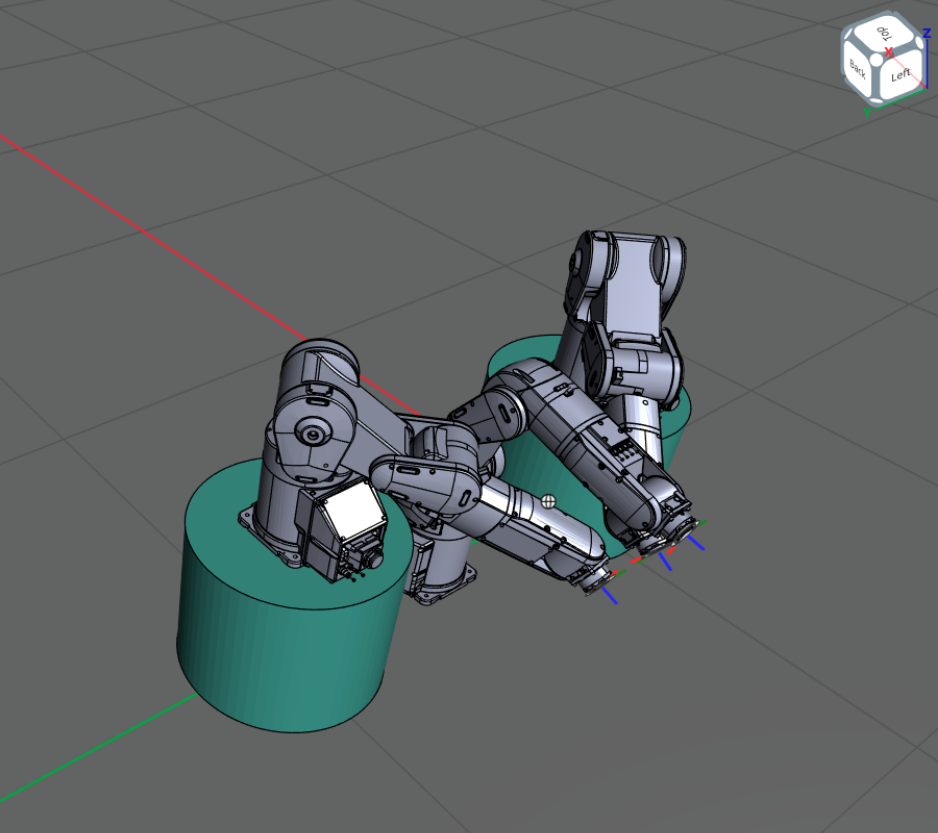}
        \caption{Narrow Passage}
    \end{subfigure}    
    \begin{subfigure}{\subfigWidth}
        \includegraphics[width=\envWidth, height=\envHeight]{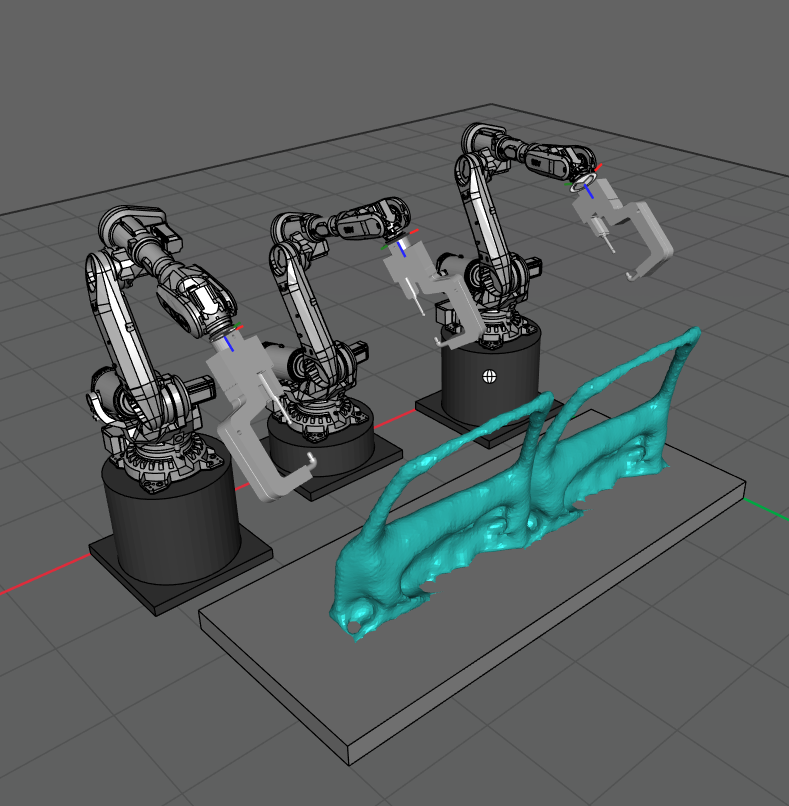}
        \caption{Welding}
    \end{subfigure}   
    \begin{subfigure}{\subfigWidth}
        \includegraphics[width=\envWidth, height=\envHeight]{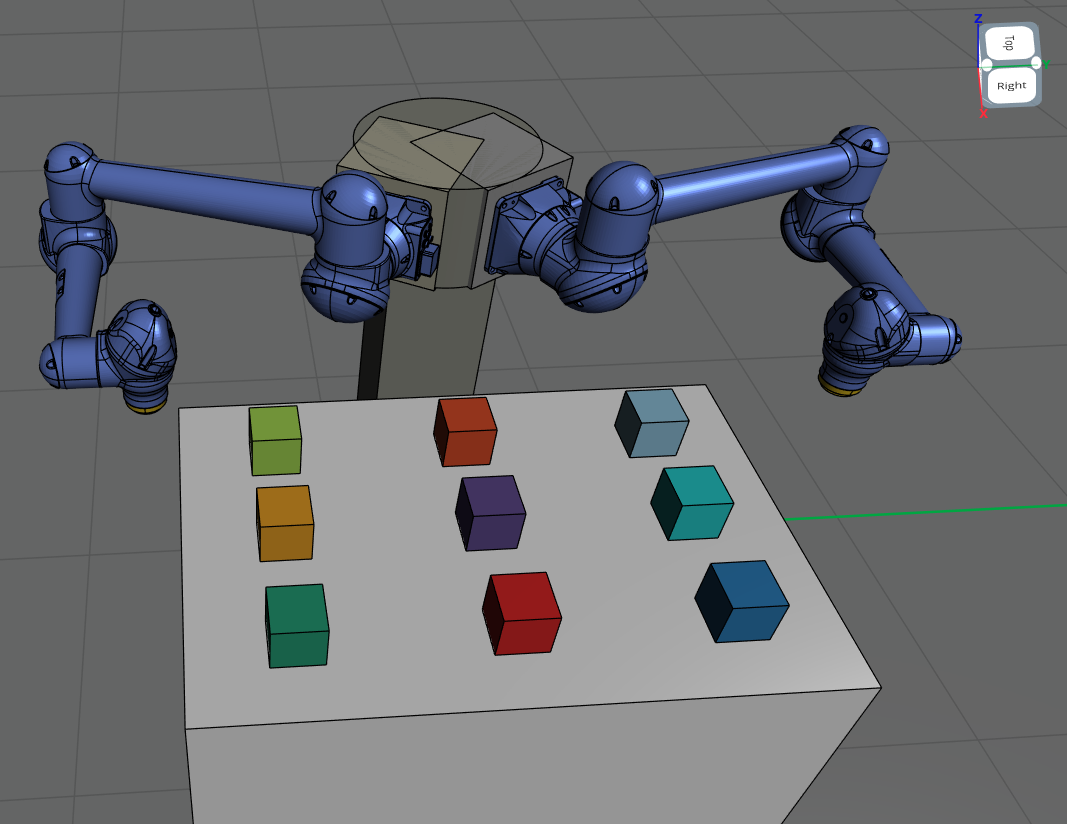}
        \caption{Pick and Place}
    \end{subfigure}
    \caption{Multi-robot planning scenarios. In each scenario, the goal for each robot is to find motion plans to reach a set of target locations while avoiding all other robots in the scene. We use robots from different manufacturers like Kawasaki, ABB, Mitsubishi, and Yaskawa.}
    \label{fig:scenarios}
\end{figure*}

%% file: images/evaluations/figure_evaluation_table.tex
\newcolumntype{T}{>{\raggedright}m{.2\linewidth}}
\newcolumntype{Y}{>{\centering}m{.18\linewidth}}
\renewcommand\tabularxcolumn[1]{m{#1}}
\begin{table}
\def\fastdrrt{Fast-dRRT*}
\centering
\begin{tabularx}{\linewidth}{|TY|XXXXX|} 
\hline
Property & Planner 
& \rotatebox{90}{Deadlock Aisles }
& \rotatebox{90}{Deadlock Table}
& \rotatebox{90}{Narrow Passage}
& \rotatebox{90}{Welding}
& \rotatebox{90}{Pick and Place}
\\
\hline
\multirow{2}{*}{Mean Time 1st} & dRRT* & $27$ & $216$ & $923$ & $42$ & $300$\\
& \fastdrrt & $\mathbf{21}$ & $\mathbf{78}$ & $\mathbf{49}$ & $\mathbf{28}$ & $\mathbf{255}$\\
\hline
\multirow{2}{*}{Worst Time 1st} & dRRT* & $32$ & $271$ & $1147$ & $163$ & $1175$\\
& \fastdrrt & $\mathbf{24}$ & $\mathbf{97}$ & $\mathbf{70}$ & $\mathbf{108}$  & $\mathbf{888}$\\
\hline
\multirow{2}{*}{Mean cost 1st} & dRRT* & $\mathbf{99}$ & $\mathbf{147}$ & $\mathbf{156}$ & $\mathbf{63}$ & $\mathbf{55}$\\
& \fastdrrt & $116$ & $161$ & $212$ &  $77$ &  $66$\\
\hline
\multirow{2}{*}{Mean cost $100$ms} & dRRT* & $\mathbf{83}$ & $\mathbf{120}$ & $-$ & $\mathbf{44}$ & $\mathbf{52}$\\
& \fastdrrt & $104$ & $151$ & $\mathbf{168}$ & $58$ & $59$ \\
\hline
\end{tabularx}

\caption{Comparison of results on the five scenarios from Fig.~\ref{fig:scenarios}. The times are given in ms (milliseconds), while the cost is the path length in configuration space.
\label{table:results}}
\end{table}

%% file: images/evaluations/figure_evaluation_time.tex
\def\gWidth{0.19\textwidth}
\begin{figure*}
    \centering
    \begin{subfigure}{\gWidth}
        \includegraphics[width=\textwidth]{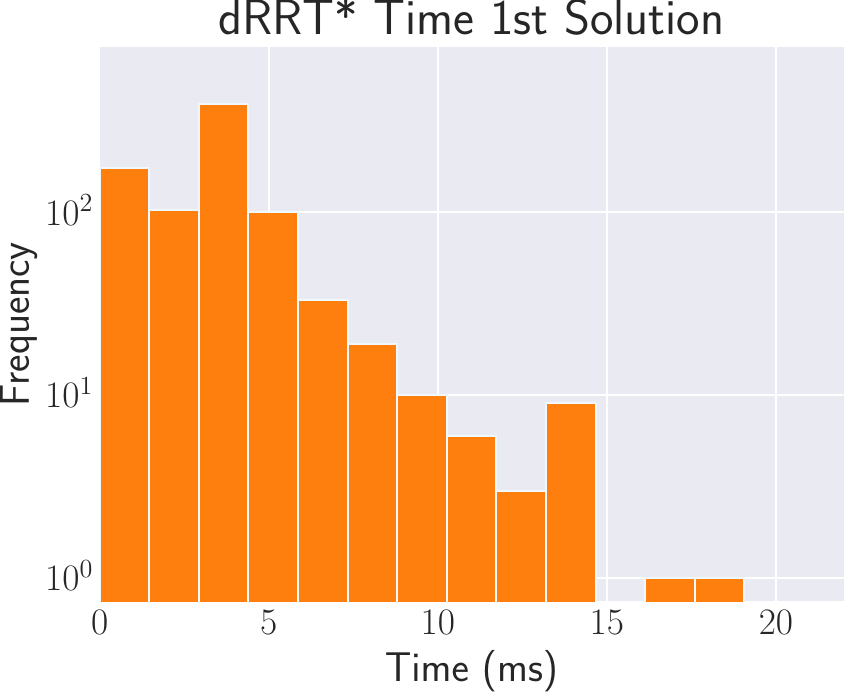}\\
        \includegraphics[width=\textwidth]{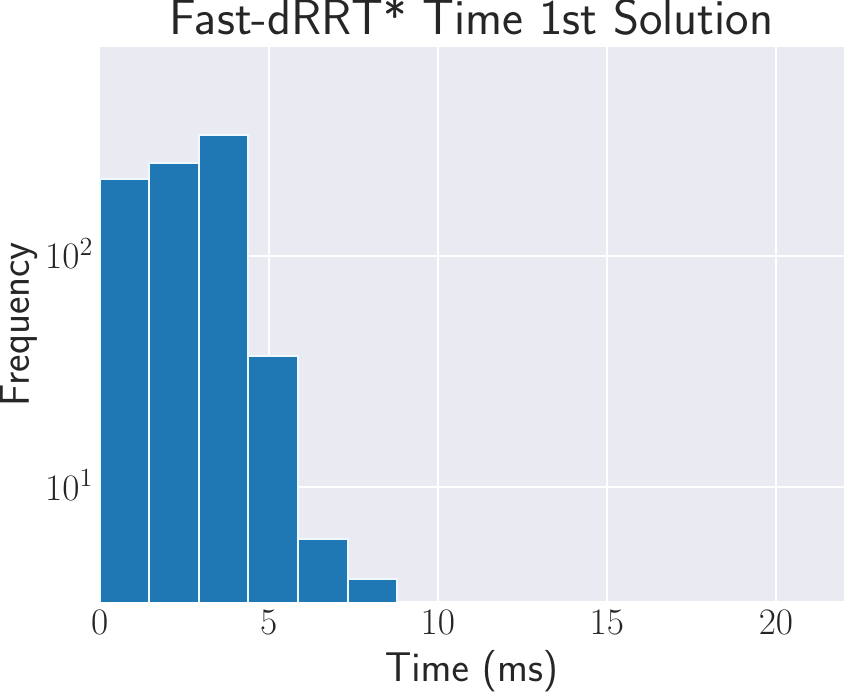}
        \caption{Deadlock Aisle}
    \end{subfigure}
    \begin{subfigure}{\gWidth}
    \includegraphics[width=\textwidth]{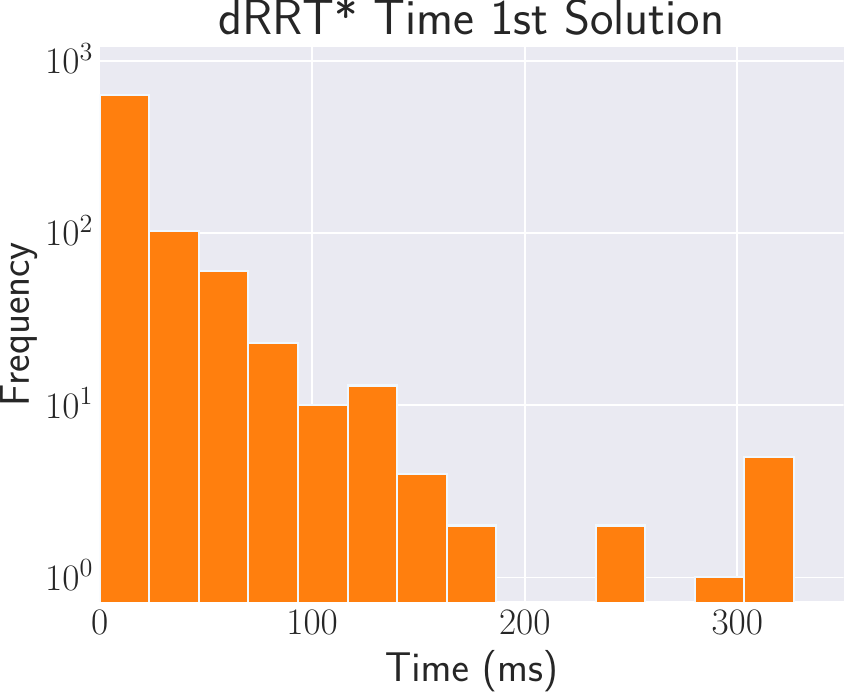}\\
    \includegraphics[width=\textwidth]{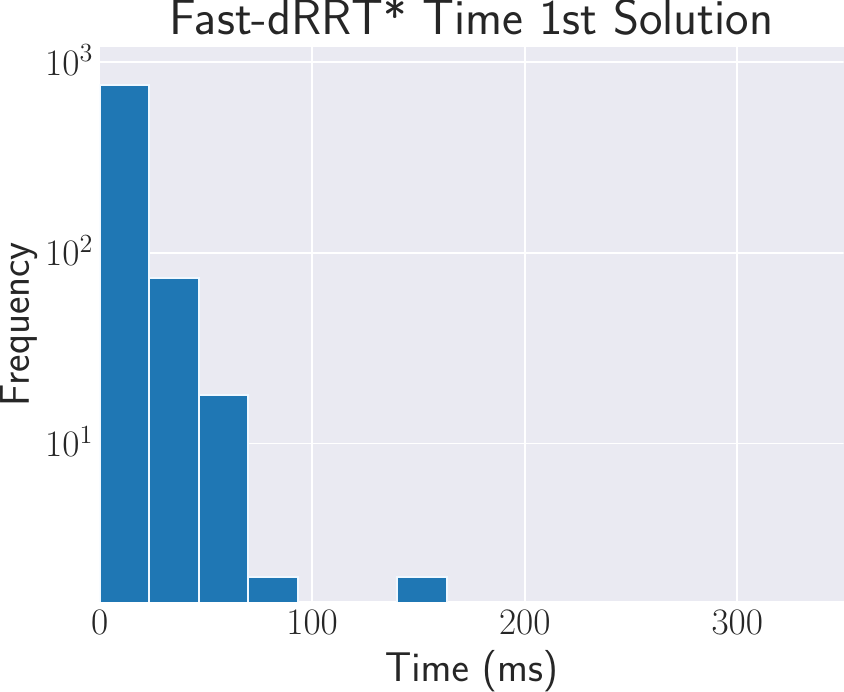}
        \caption{Deadlock Table}
    \end{subfigure}
    \begin{subfigure}{\gWidth}
    \includegraphics[width=\textwidth]{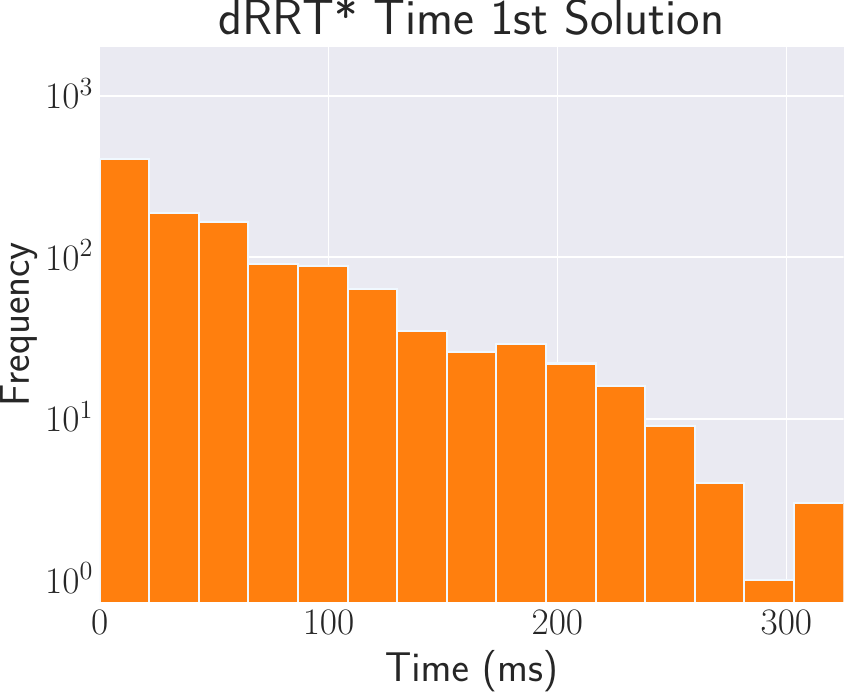}
    \includegraphics[width=\textwidth]{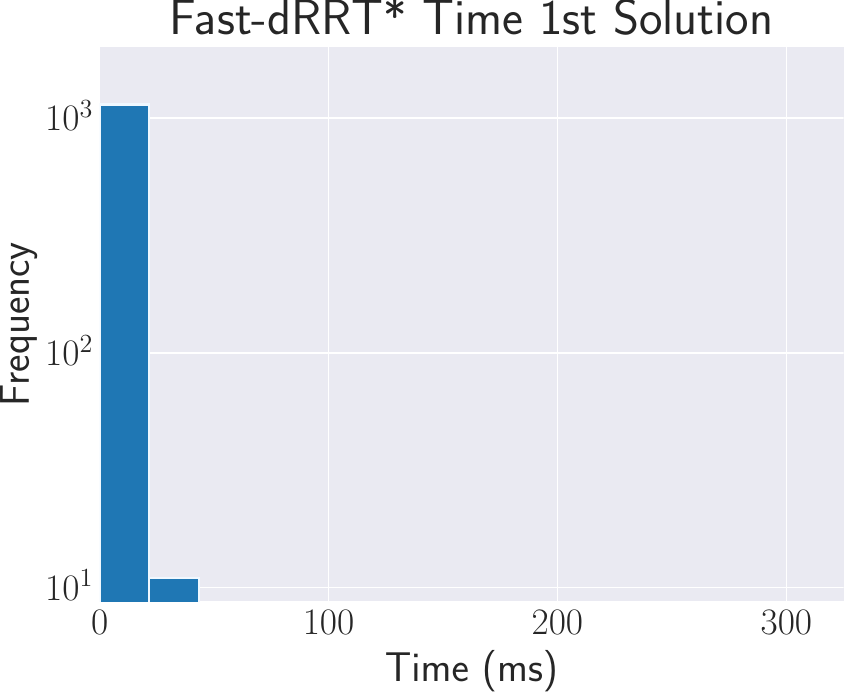}
        \caption{Narrow Passage}
    \end{subfigure}
    \begin{subfigure}{\gWidth}
    \includegraphics[width=\textwidth]{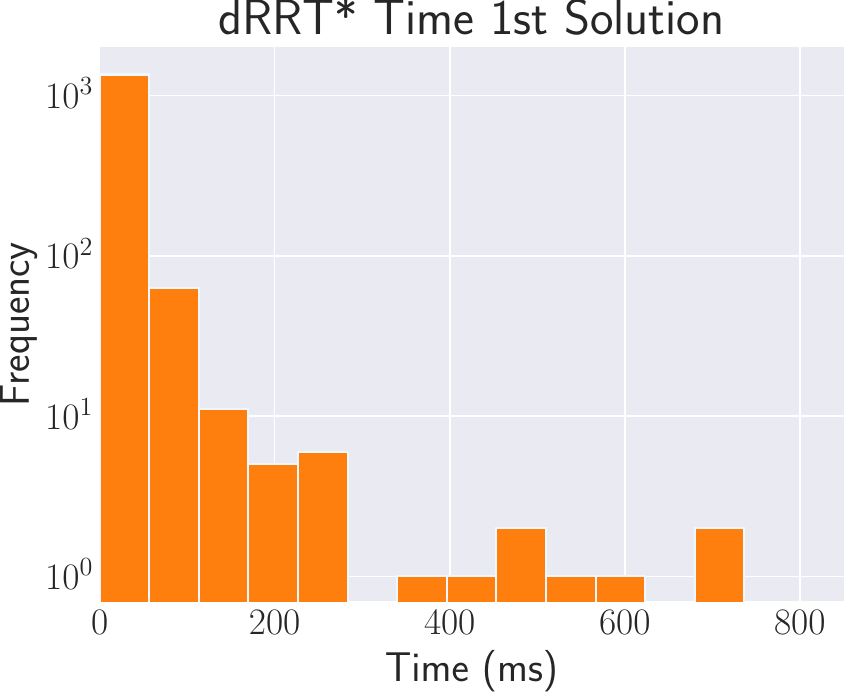}
    \includegraphics[width=\textwidth]{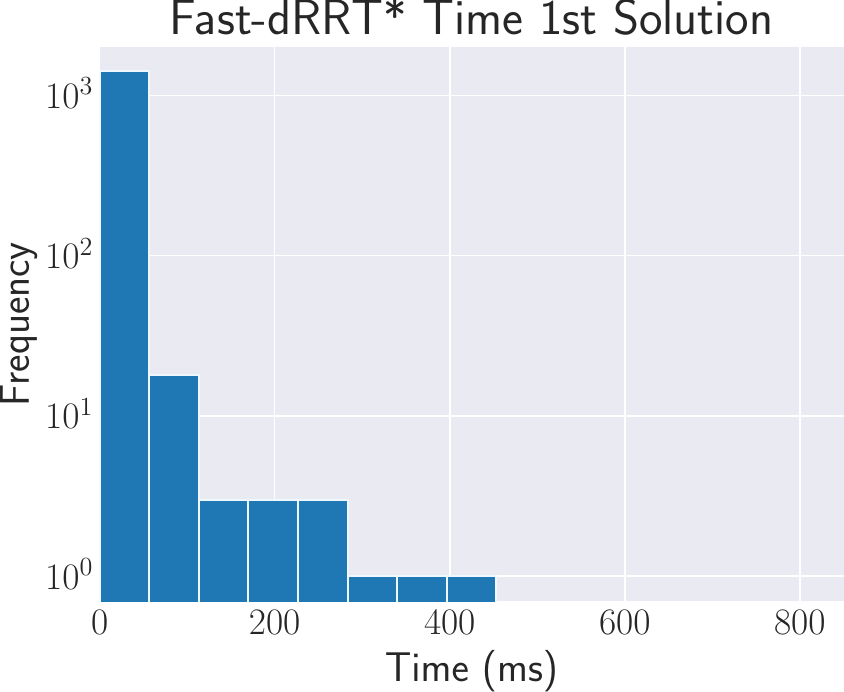}
        \caption{Welding}
    \end{subfigure}
    \begin{subfigure}{\gWidth}
    \includegraphics[width=\textwidth]{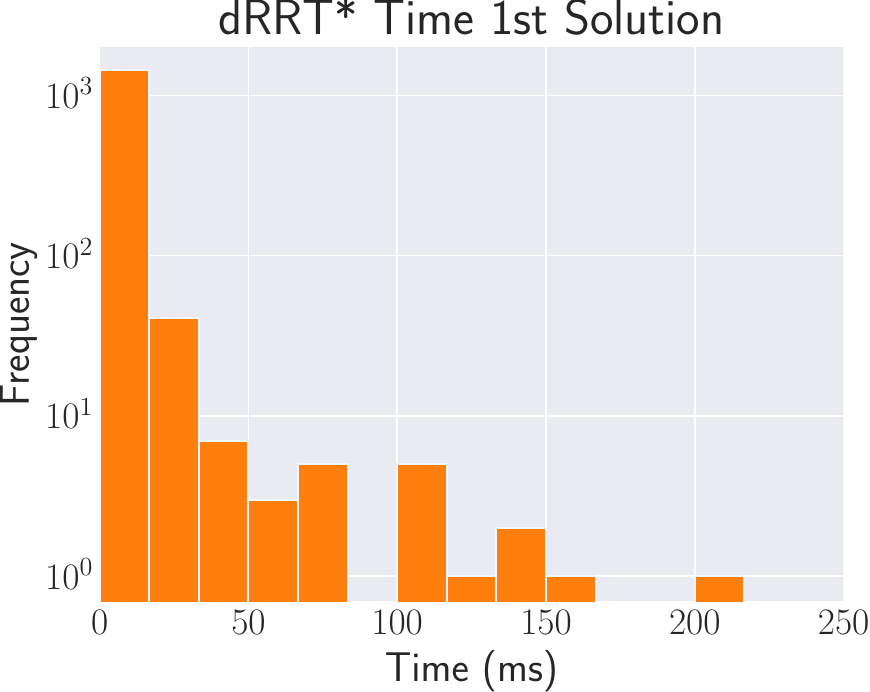}\\
    \includegraphics[width=\textwidth]{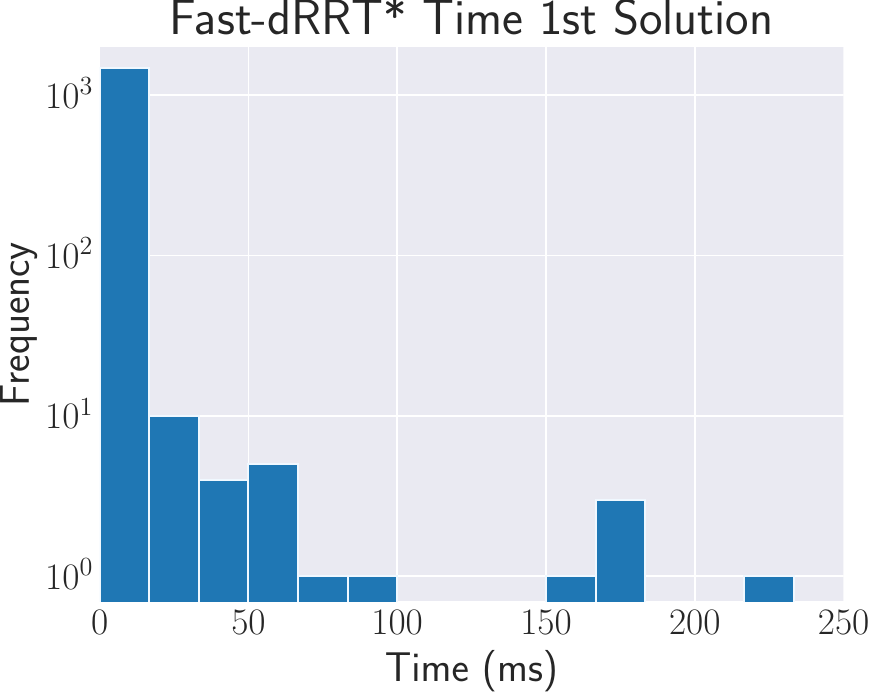}
        \caption{Pick and Place}
    \end{subfigure}
    \caption{Comparison on planning time for all five scenarios from Fig.~\ref{fig:scenarios}. \textbf{Upper row:} Histogram of dRRT* results, showing how often a result falls into a certain time slice. $x$-axis has been adjusted for better visibility. \textbf{Lower row:} Corresponding histogram of Fast-dRRT* results.}
    \label{fig:time_evaluation}
\end{figure*}

%% file: images/evaluations/figure_evaluation_cost.tex
\def\gWidth{0.19\textwidth}
\begin{figure*}
    \centering
    \begin{subfigure}{\gWidth}
        \includegraphics[width=\textwidth]{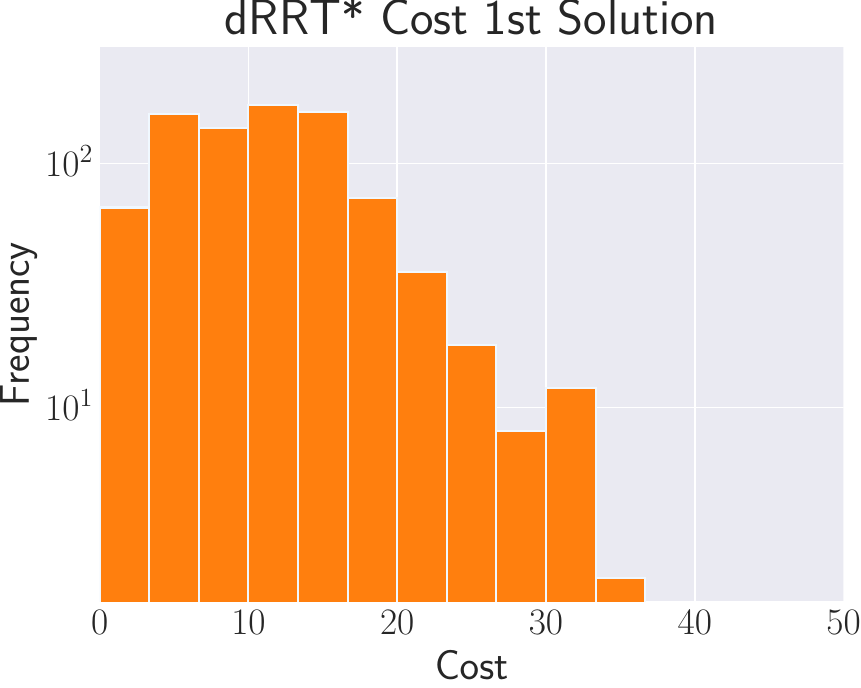}
        \includegraphics[width=\textwidth]{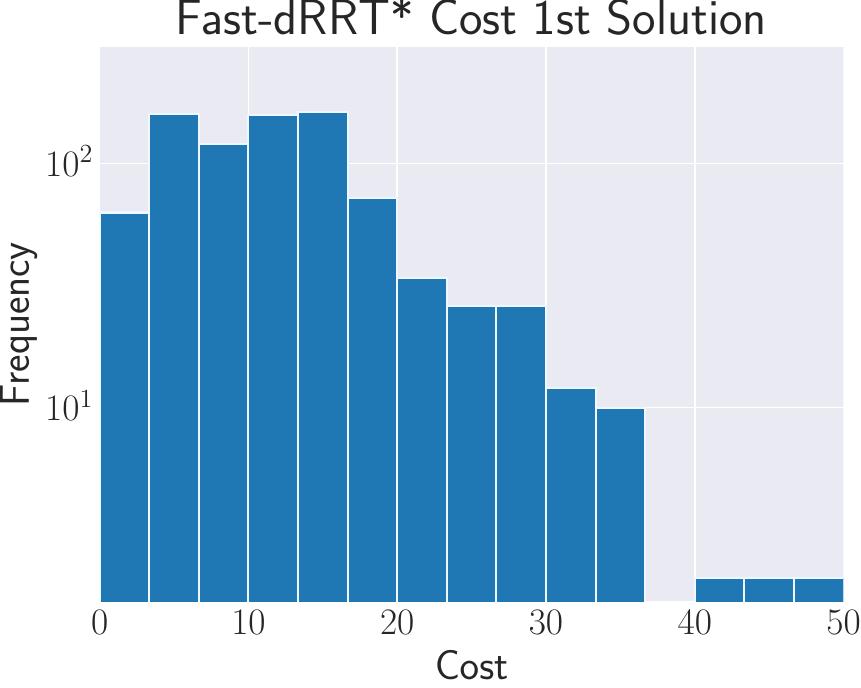}
        \caption{Deadlock Table}
    \end{subfigure}
    \begin{subfigure}{\gWidth}
        \includegraphics[width=\textwidth]{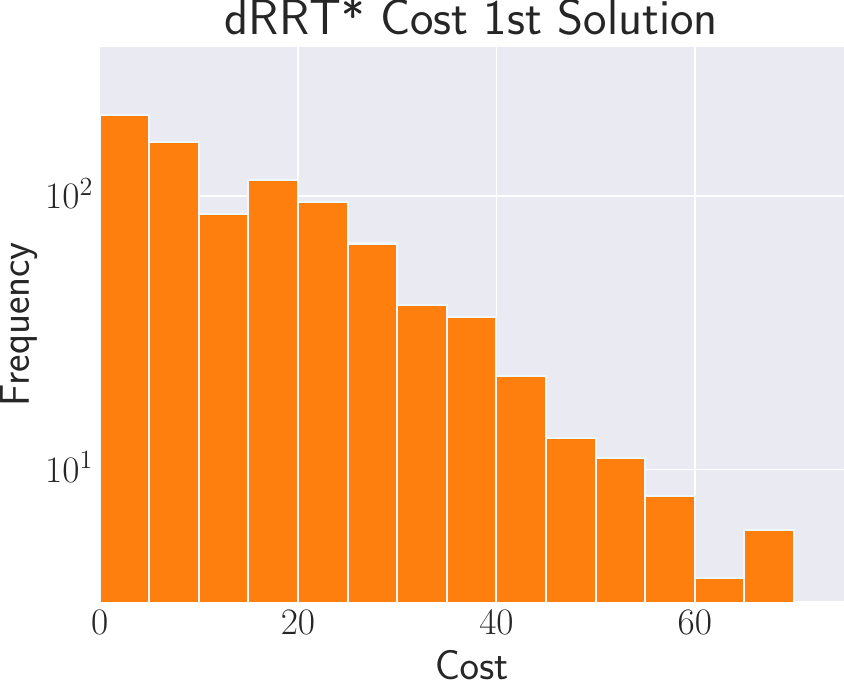}
        \includegraphics[width=\textwidth]{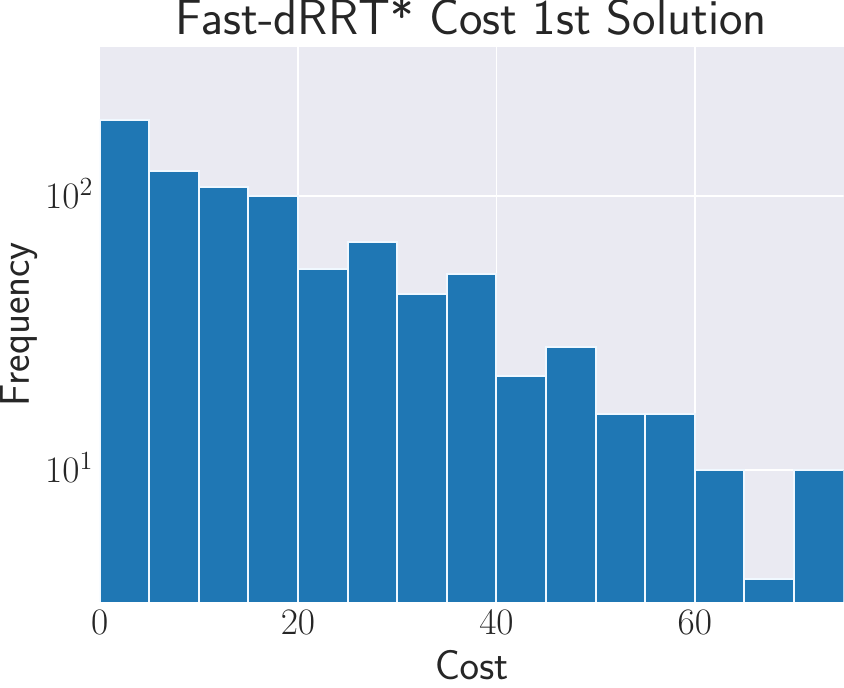}
        \caption{Narrow Passage}
    \end{subfigure}
    \begin{subfigure}{\gWidth}
        \includegraphics[width=\textwidth]{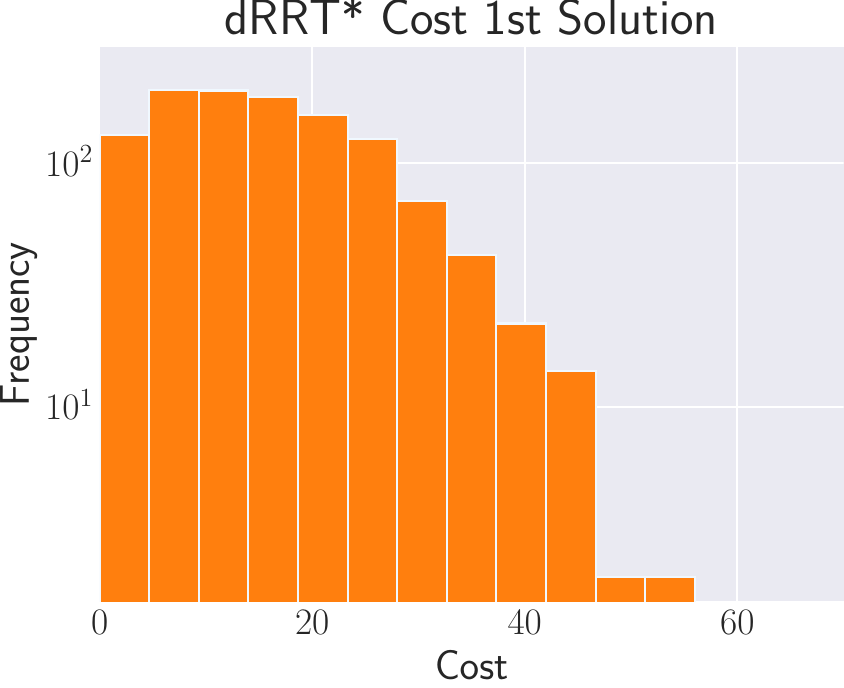}
        \includegraphics[width=\textwidth]{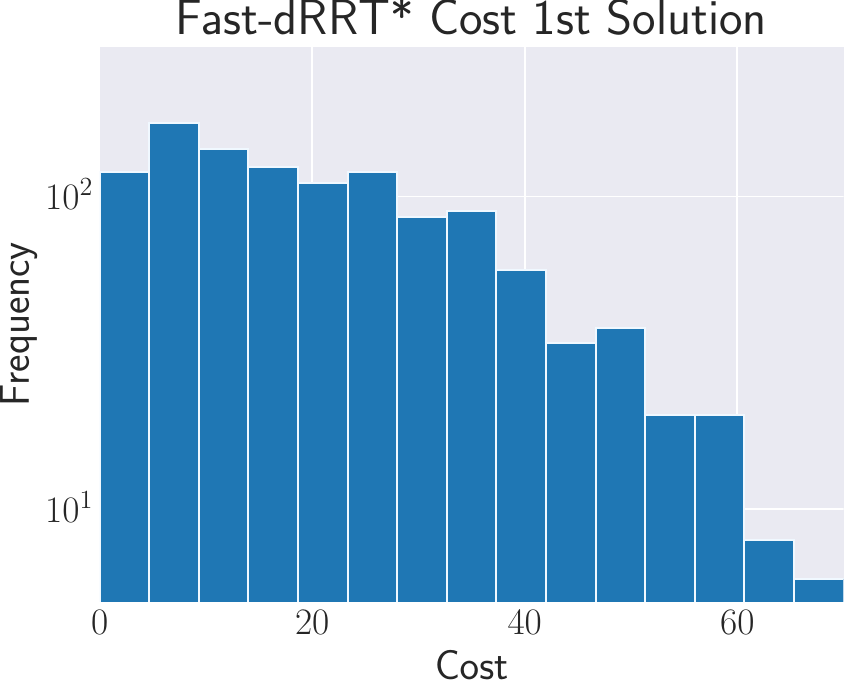}
        \caption{Deadlock Aisle}
    \end{subfigure}
    \begin{subfigure}{\gWidth}
        \includegraphics[width=\textwidth]{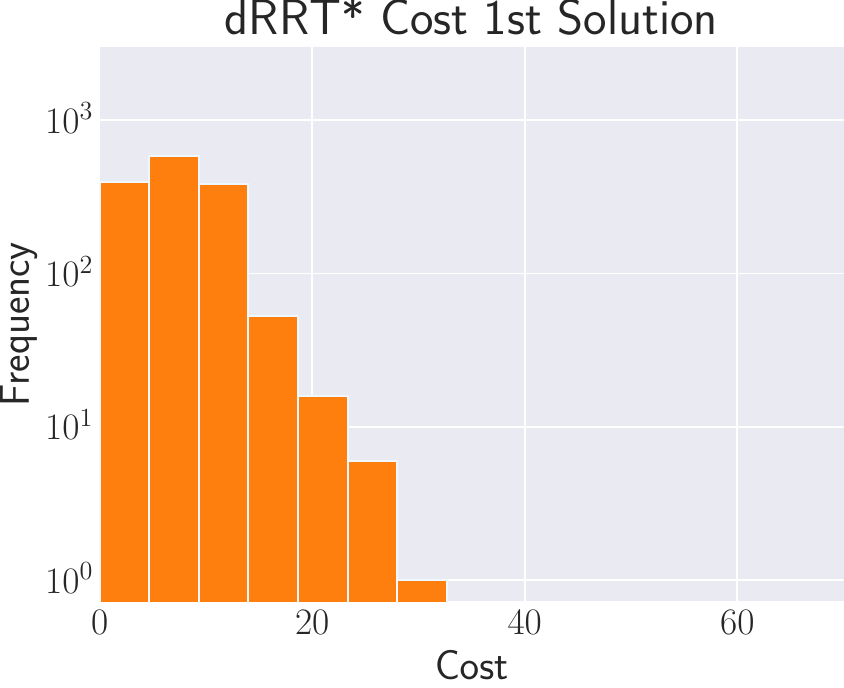}\\
        \includegraphics[width=\textwidth]{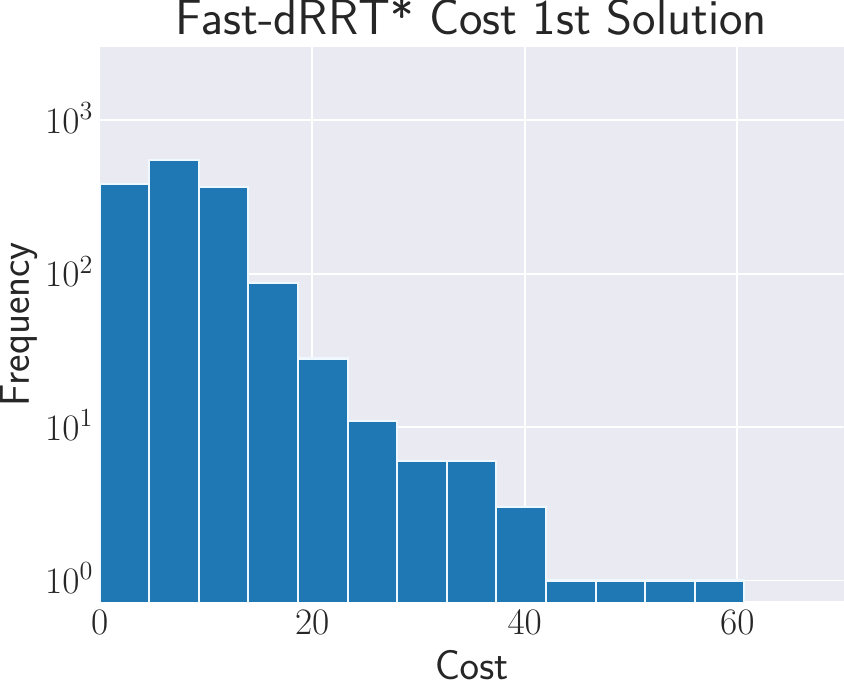}
        \caption{Welding}
    \end{subfigure}
    \begin{subfigure}{\gWidth}
        \includegraphics[width=\textwidth]{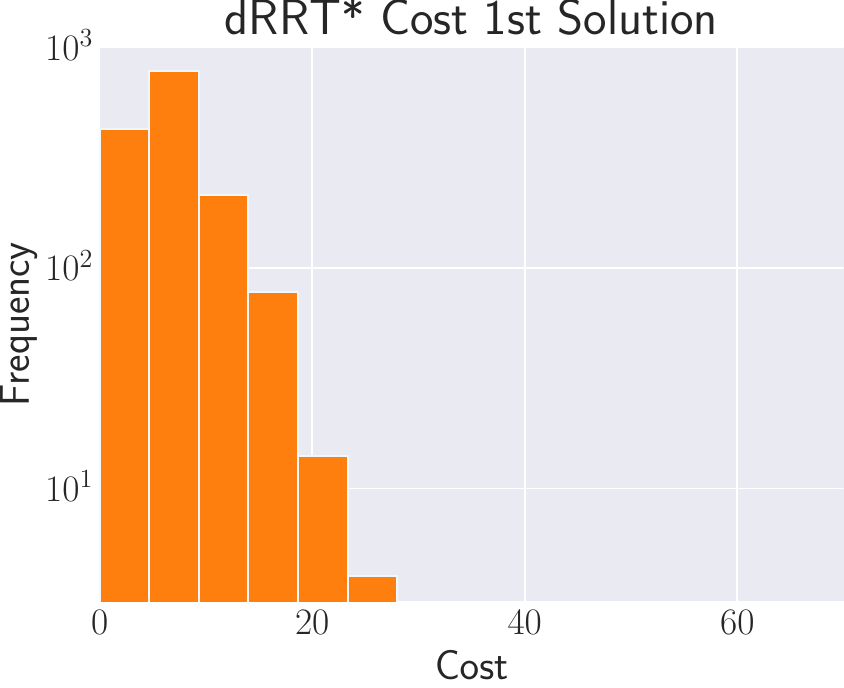}\\
        \includegraphics[width=\textwidth]{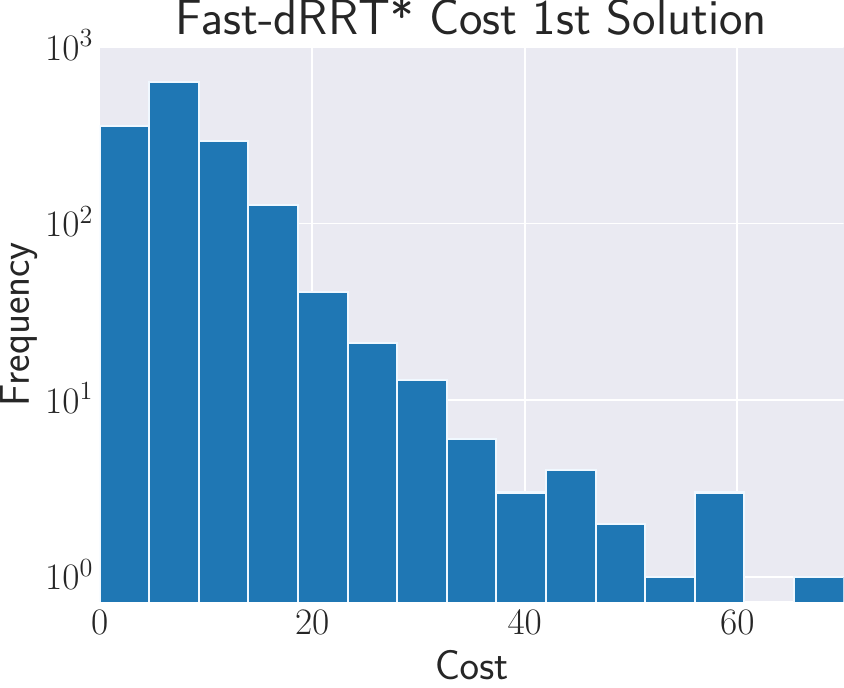}
        \caption{Pick and Place}
    \end{subfigure}
    \caption{Cost comparison on all five scenarios from Fig.~\ref{fig:scenarios}. \textbf{Upper row:} Histogram of dRRT* cost results after a first solution is found. \textbf{Lower row:} Corresponding histogram for Fast-dRRT*.
    \label{fig:cost_evaluation}}
\end{figure*}

%% file: src/06_conclusion.tex
\section{Conclusion\label{sec:conclusions}}

We presented Fast-dRRT*, which adapts dRRT*~\cite{shome_drrt_2020} to reduce planning time for industrial automation scenarios. 
Fast-dRRT* can exploit pre-computed swept volumes, is able to solve partially specified multi-robot problems, and improves upon computationally expensive parts of dRRT*.

In our experiments, we compared Fast-dRRT* against dRRT* on realistic industrial scenarios, whereby both planner used pre-computed swept volumes, and deadlock avoidance. 
Both dRRT* and Fast-dRRT* were able to solve most challenges within the time limit, whereby Fast-dRRT* consistently outperformed dRRT* in terms of the time required to find an initial solution. 
However, its convergence time to optimal solutions is higher than that of dRRT*, and the initial solutions it finds might have higher cost. 
Additionally, both planners are adapted to handle partially specified multi-robot problems by moving blocking robots out of the way, thus making them more versatile and capable of solving a broader range of scenarios compared to the original dRRT*~\cite{shome_drrt_2020}.

We believe there are several ways to improve Fast-dRRT*. 
First, we could reduce the convergence time of Fast-dRRT* by integrating optimization methods during roadmap construction~\cite{Kamat2022IROS}. 
Another important extension would be the inclusion of end-effector constraints, to either keep the end-effector in certain orientations, or to keep it inside a specific workspace. 
Also, Fast-dRRT* could be improved using sparse roadmaps~\cite{dobson_2014, Orthey2021ICRA} which selectively sparsify individual roadmaps in non-important regions of the state space, thereby speeding up the overall runtime.

Despite possible improvements, the current results indicate that Fast-dRRT* algorithm is an efficient method to deal with real-world industrial applications. This makes Fast-dRRT* an important contribution to reduce production costs and to increase productivity in automated industrial manufacturing.